# AI for pRedicting Exacerbations in KIDs with aSthma (AIRE-KIDS)

Hui-Lee Ooi[1,2], Nicholas Mitsakakis[1], Margerie Huet Dastarac[1,2], Roger Zemek[1,2], Amy C. Plint[1,2], Jeff Gilchrist[1], Khaled El Emam[1,2]*, Dhenuka Radhakrishnan[1,2]*

[1] CHEO Research Institute, Ottawa, Ontario, Canada
[2] Faculty of Medicine, University of Ottawa, Ottawa, Ontario, Canada

*Senior authors

**Corresponding Author:**
Dhenuka Radhakrishnan
Children's Hospital of Eastern Ontario
401 Smyth Road, Ottawa, ON, K1H 8L1
Canada
Email: dradhakrishnan@cheo.on.ca

## Abstract

**Background:** Recurrent exacerbations remain a common yet preventable outcome for many children with asthma. Machine learning (ML) algorithms using electronic medical records (EMR) could allow accurate identification of children at risk for exacerbations and facilitate referral for preventative comprehensive care to avoid this morbidity.

**Objectives:** We developed ML algorithms to predict repeat severe exacerbations (i.e. asthma-related ED visits or future hospital admissions) for children with a prior asthma ED visit at a tertiary care children's hospital.

**Methods:** We used retrospective pre-COVID19 (Feb 2017 – Feb 2019, N=2716) Epic EMR data from the Children's Hospital of Eastern Ontario (CHEO) linked with environmental pollutant exposure and neighbourhood marginalization information to train various ML models. We used boosted trees (LGBM, XGB) and 3 open-source large language model (LLM) approaches (DistilGPT2, Llama 3.2 1B and Llama-8b-UltraMedical). Models were tuned and calibrated then validated in a second retrospective post-COVID19 dataset (Jul 2022 – Apr 2023, N=1237) from CHEO. Models were compared using the area under the curve (AUC) and F1 scores, with SHAP values used to determine the most predictive features.

**Results:** The LGBM ML model performed best with the most predictive features in the final AIRE-KIDS$_{ED}$ model including prior asthma ED visit, the Canadian triage acuity scale, medical complexity, food allergy, prior ED visits for non-asthma respiratory diagnoses, and age for an AUC of 0.712, and F1 score of 0.51. This is a nontrivial improvement over the current decision rule which has F1=0.334. While the most predictive features in the AIRE-KIDS$_{HOSP}$ model included medical complexity, prior asthma ED visit, average wait time in the ED, the pediatric respiratory assessment measure score at triage and food allergy.

**Conclusions:** The AIRE-KIDS models accurately predict repeat asthma-related ED visits or hospitalizations within one year in children with a previous asthma ED visit. Despite the popularity of LLM's, the LGBM approach yielded the best performing models. Future clinical deployment of AIRE-KIDS from the ED-setting as a decision support tool has the potential to transform how we enable equitable access to comprehensive preventative asthma care for the most vulnerable children directly from the ED.

# 1.    Introduction

Asthma is one of the most prevalent chronic childhood diseases and a leading cause of repeated emergency department (ED) visits and hospitalization for the management of severe exacerbations. Up to 25% of children with an asthma ED visit will have a repeat visit in the following year, despite numerous evidence-based treatments available to optimize asthma control [1]. For example, comprehensive asthma self-management education is a cornerstone for improving asthma control and reducing exacerbations; however, capacity constraints limit the feasibility of providing this intervention for all children with asthma, especially in the ED setting.  With current shortages in primary care, this type of preventative intervention is also frequently missed in the ambulatory setting. A novel approach is to target this or other resource-intensive treatment approaches, including specialist care, for the patients or families who are most at risk for repeat asthma ED visits directly from the ED setting. Currently, there is a lack of guidance or agreed consensus on how to identify this vulnerable group of children who are most at risk for repeat severe asthma exacerbations. Given the increase in the use of electronic medical records (EMR) at many hospitals and the consequent availability of large amounts of routinely collected clinical data available for children with asthma healthcare encounters, there may be an opportunity to apply machine learning models to address this gap.

While there is a larger literature on predicting repeat asthma exacerbations in adults, the number of studies in children is limited.  These previous pediatric studies are heterogeneous in how asthma exacerbations were defined and generally used traditional biostatistical approaches to determine the importance of various predictors for this outcome [2–8]. The definition of severe asthma exacerbation [9] in prior studies ranges from requiring oral corticosteroids (OCS), hospitalization, asthma related emergency visits or even death [5,7,8,10–20]. Among the studies that defined severe exacerbation as a repeat ED visit or hospitalization, the observation windows from the initial visit varies from 15 days to one year [5,11,13–16,21,22].  However, as asthma is a seasonal disease, with exacerbation risk dependent on triggers that may be more or less prevalent at different times of the year, a one-year observation for outcomes is the most relevant timeframe for long term management and prevention of exacerbations. Previously reported predictors of repeat asthma ED visit or hospitalization over one year have included prior acute care visits, patient-level characteristics obtained from the patient chart or EMR and environmental information [3,7,11–14,23].

Logistic regression and machine learning (ML) models, such as tree-based approaches, ensemble methods as well as deep learning techniques have been applied to predict asthma exacerbation [6,7,13–

[18,24,25], including (MLP) [5,14,17,26], recurrent neural networks [27–29] and long short-term memory models (LSTM) [30]. Recently, large language models (LLMs) [31–34] have gained momentum in the prediction of health outcomes but have not yet been applied to predicting asthma exacerbations. The benefit of LLMs over previous ML techniques lies in their improved potential for application into clinical care as these models may be more transferable across sites with minimal retraining since they are already pre-trained on a large amount of general and medical information. Despite the considerable potential of leveraging these large data sources, a recent article concluded that there remain concerns that LLMs cannot perform at a comparable level to ML models [35]. This type of comparison has not been performed on asthma exacerbation prediction, hence the importance of evaluating LLM performance against traditional ML models in this specific clinical context. Multiple LLM approaches have been investigated and implemented in the literature, and we can categorize them into open-source model fine-tuning [31,36,37] and larger proprietary model in-context learning [32,33,38]. In clinical settings open-source models where inference can be performed locally are preferred due to strict privacy rules that limit the ability of transferring personal health information to external sites that do not meet regulatory security requirements.

The aim of the current study was to evaluate several machine learning (ML) approaches, including 3 different LLMs, to best identify children with an asthma ED visit who are at risk for repeat asthma ED visits or hospital admissions over a one-year timeframe, using routinely collected information in the Epic EMR at a tertiary care pediatric centre in Ontario, Canada. Future implementation of an accurate ML algorithm into clinical workflows in the ED would allow streamlined preventative care for children most at risk for repeat asthma ED visits or hospitalizations.

## 2. Methods
### 2.1 Study Design

In this study, we used retrospective Epic EMR data to train ML models for predicting future severe asthma exacerbations among patients treated for an asthma exacerbation in the ED at the Children's Hospital of Eastern Ontario (CHEO). These "high risk" patients would be referred to the CHEO asthma program, which is a stepped-care program that triages patients to receive comprehensive asthma education by a certified asthma educator and/or referral to an asthma specialist at the discretion of the asthma educator [41].

A future severe asthma exacerbation was defined as an ED re-visit or hospitalization with asthma as the most responsible diagnosis within one year of the initial asthma ED visit. We used two different retrospective datasets for training and validating models. The training dataset (pre-COVID-19) accrued patients over 2 years between February 2017 to March 2019, with outcomes observed up to March 2020. Similarly, we created a post-COVID validation dataset with an accrual period from July 2022 to April 2023, with outcomes observed up to April 2024.

We defined an 'index visit' as the first ED visit with a diagnosis of asthma (i.e. International Classification of Diseases and Related Health Problems version 10 (ICD-10) codes J45, J46) within the defined accrual period (see Figure 1). Patients with an index visit for asthma during the peak COVID19 period (March 2020-June 2022) were not included in either cohort due to a significantly lower frequency of asthma exacerbations observed following provincial infection control measures implemented in Ontario [42].

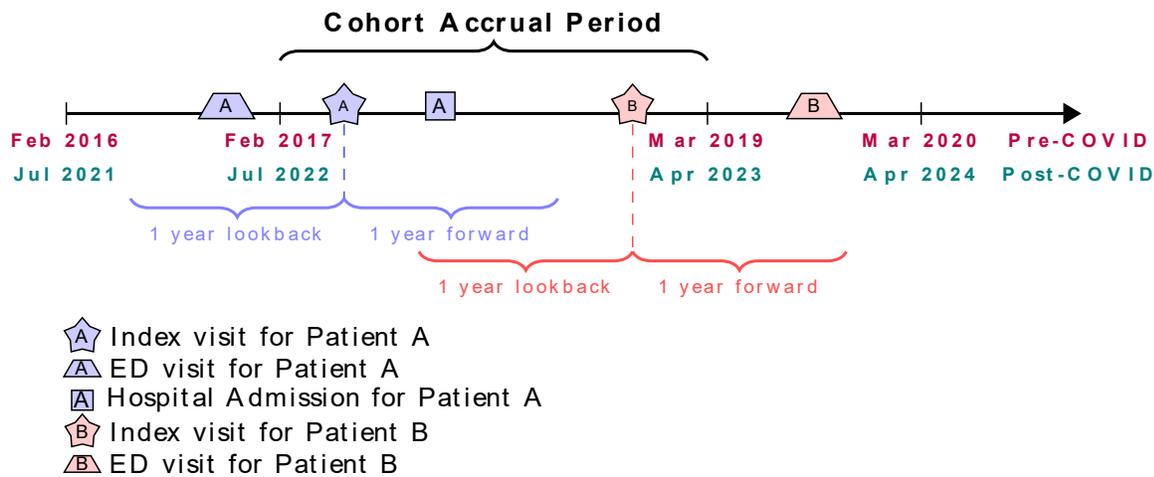

**Figure 1:** Study timelines and definition of accrual period with two examples for patients A and B. Note for patient A there is an ED visit outside the accrual period which would be considered as a predictive feature and not the index event. ED = emergency department

The ML models were compared to the current standard of care at CHEO where a best practice alert has currently been implemented in the EMR to alert ED clinicians to children who would be eligible for referral to the CHEO asthma program based on prespecified criteria that have previously been shown to correlate with a higher risk for repeat asthma hospitalization [43] (see Figure 2).

This study is reported according to Consolidated Reporting of Machine Learning Studies (CREMLS) reporting guidelines [44]. The reporting checklist is included in the Supplementary materials.

> **CHEO Best Practice Alert for referral to the CHEO asthma program**
>
> - patient returning to ED with history of prior asthma ED visit in previous 12 months AND:
> - a PRAM score of ≥ 3 at triage during current visit OR
> - received systemic corticosteroids during the current visit

ED = emergency department, PRAM = Pediatric Respiratory Assessment Measure, a previously validated tool to indicate the severity of respiratory symptoms during an asthma exacerbation and informs algorithm-based asthma management in the ED. Score 1-3 = mild, 4-7 = moderate, 8-12 = severe [45]

**Figure 2:** Current (2025) CHEO best practice rule to identify patients who would qualify for referral to the CHEO asthma program where they would receive comprehensive asthma education or specialist referral due to a presumed higher risk of future asthma hospitalization.

## 2.2 Predictors/ Features

We considered all features that are available in the CHEO Epic EMR and that have clinical plausibility or have been previously shown to be associated with the risk of repeat asthma exacerbation [46] (and also see the summary of previous studies in the Supplementary materials). A total of 68 clinical features were extracted from the EMR and further linked to air pollution and neighborhood marginalization data.

Among the variables extracted from the EMR, we included measures of prior health care utilization, including previous asthma-related ED visits or admissions, as well as prior-non asthma respiratory ED visits or admissions within one year prior to the index visit.

We captured operational variables related to each of these healthcare visits including factors reflecting the volume of patients and wait time in the ED such as day and month of the index visit, time from arrival to triage or disposition, the hourly average number of patients present in the ED as well as the average wait time during each patient's visit. Demographic features included age during the index visit, sex, whether the patient has a primary care physician, the distance of the hospital from the postal code of the patient's home and the province of the patient's home as patients from both Ontario and the neighboring province of Quebec are treated at CHEO.

We additionally included features reflecting the severity of the asthma presentation at the index visit such as the first oxygen saturation recording, timing of order and administration of oral corticosteroids, the Pediatric Respiratory Assessment Measure (PRAM) score at triage, and the Canadian Triage Acuity Score (CTAS). PRAM is a previously validated tool to indicate the severity of respiratory symptoms during an asthma exacerbation and informs algorithm based asthma management in the ED (i.e. Score 1-3 = mild, 4-7 = moderate, 8-12 = severe) [45], and CTAS is a triage scale ranging from one to five that is used for all ED patients (level 1 = highest) to identify illness acuity and the time frame for assessment [47].

Among the clinical variables we included were the presence of food and environmental allergens, lab test results such as serum eosinophils count and immunoglobulin E levels, medical complexity as reflected in prior visits to subspecialty clinics, and the presence of specific comorbidities that are known to correlate with asthma severity.

For a subset of the cohort, that is patients residing in Ontario, we additionally included features related to air pollution and neighbourhood marginalization quintile. Specifically, the Air Quality Health Index (AQHI), a composite measure of air pollution that is shown to correlate with health and emergency department use [48], as well as mean and highest Nitrogen Dioxide, Ozone, and $PM_{2.5}$ readings were considered at different intervals (i.e. 1 day, 2 days and 3 days prior to the index visit) as measured at specified postal codes in the catchment area of CHEO. The residence postal code of each patient was matched to the nearest postal code where air pollutants were measured, to allow linkage of these features. Similarly, we used residence postal code to link with each child's neighbourhood marginalization quintile as defined by the Ontario marginalization (ON-Marg) index, a validated tool that classifies marginalization in 4 domains: households and dwellings – a reflection of family and neighbourhood stability, material resources, age and labour force – a measure of disability and dependence, and racialized and newcomer populations.

Further details of these and additional predictor variables are included in Supplementary materials.

## 2.3     Data and Study Population

Data related to clinical characteristics and hospital, outpatient or ED visits were collected from the hospital Epic EMR system while social and environmental variables were obtained from public data sources [49]. Patients were included in the cohort if they had an asthma diagnosis (ICD code J45, J46) at the index encounter during the accrual period. Patients were excluded if they had attended a respirology appointment or received comprehensive asthma education prior to the index visit, or if they were admitted to hospital during the same encounter (i.e., on the same day) as the index asthma ED visit.

## 2.4     ML Model Outcomes

Our two main outcomes of interest were 1) repeat asthma ED visit (which would include future asthma hospital admissions that occur subsequent to an ED visit) within 1 year of the index ED visit, and 2) future asthma hospital admission within 1 year of the index ED visit. Of note, children or families who participate in the CHEO asthma program and receive comprehensive asthma education or are seen by an asthma specialist may have a reduced risk of repeat severe asthma exacerbation. This potential source of 'contamination' of the dataset was accounted for by defining both the repeat ED visit outcome or hospitalization outcome as: 1) patients who returned to the ED for asthma, or who received asthma education or attended a respirology clinic visit, all within one year of the index asthma ED visit, and 2) patients who were hospitalized for asthma, or who received asthma education or attended a respirology clinic visit, all within one year of the index asthma ED visit.

## 2.5 ML Models Evaluated

In this study, we considered two types of boosted decision trees and three open-source LLMs with finetuning. The two types of boosted trees evaluated were LGBM [50] and xgboost [51]. Tree-based models are the most common type of ML prognostic methods used in clinical research [52–57], they perform better than linear models, such as logistic regression [53–57], and better than deep learning on tabular datasets [58,59]. The hyperparameters of boosted tree models were tuned and optimized using Bayesian optimization with 5-fold cross-validation [60]. The range for the tuning parameters, specific to each model, was defined as recommended in the literature [61–64].

For binary classification using boosted trees, the predicted values are not true probabilities that reflect likelihood of an outcome among patients [65,66]. Therefore, these pseudo probabilities were calibrated after the fact using beta calibration, which provides optimal results for small data sets and boosted trees modeling approaches [67].

In order to use a LLM for tabular data, we applied a textual encoder [31] which transforms each tabular record into text. The LLMs, initially pretrained to generate text were fine-tuned to predict the outcome variables from the data, namely emergency department visit and admission. To address model stochasticity, all predictions were generated five times for each patient and the predicted values were averaged. Additional details about the fine-tuning process are in the Supplementary files.

Following the privacy rules at our centre, we were not able to utilize the commercial off-site LLM's for this study and were limited to those that could be executed locally. We fine-tuned three pretrained open-source LLMs available on HuggingFace [68]: DistilGPT2, Llama 3.2 1B [69] and Llama-8b-UltraMedical. The model DistilGPT2 is a lightweight LLM obtained from applying knowledge distillation [70] on GPT-2 [71], a model from OpenAI. DistilGPT2 has been used in most publications that tackle fine-tuning LLMs for tabular data [31,36,72]. Llama 3.2 1B is a more recent model released by Meta and trained on general semantic data and Llama-8b-UltraMedical is larger and has been specialized on medical data by using the UltraMedical dataset [73].

## 2.6 Evaluating ML Model Performance

Capacity within the CHEO asthma program is resource-limited and may not be available to all patients with the predicted outcome. Therefore, we wanted a model that maximizes the proportion of patients that were predicted as high risk that were actually high risk (precision), hence ensuring that the resources are used most effectively. However, a model can trivially achieve high precision by correctly predicting a single patient as high risk. Therefore it is also important to consider the proportion of high risk patients that were correctly predicted as high risk (recall). Therefore, the F1 score was deemed to be the most appropriate evaluation metric for our context, which is the harmonic mean of precision and recall.

To compute recall and precision from the predicted probability, we experimented with different cutoff thresholds used in previous studies, for example, by selecting the number of positive predictions [74], selecting a cutoff for a pre-determined specificity [6], top percentile [11,12,40], the use of a validation set [39], and lowest cost from the cost function on the training cohort [75]. However, to reflect the aim of our study which was to identify the most accurate ML model to predict repeat asthma ED visits or hospitalization to streamline referral of patients directly from the ED to the CHEO asthma program, we selected cutoffs that would maximize the F1 score.

Using 5-fold cross-validation on the training dataset, the F1 score was maximized by setting different thresholds (cutoff values at which the predicted outcome is considered positive or negative) for precision and recall, ranging from 0.05 to 0.99. The threshold that achieved the highest F1 score was averaged across the five folds and subsequently selected as the threshold of choice. This optimized threshold was then applied when running the model on the validation dataset.

Since the value of the F1 score is affected by prevalence, its proper interpretation requires comparison against one or more baselines. We established three baselines: (a) the 'CHEO best practice' rule that is currently used in the ED to identify high risk patients (i.e. current standard), (b) a naïve prediction rule that predicts that everyone is high risk, and for the F1 score this gives the largest possible naïve value, and (c) a random prediction of risk status based on a 0.5 probability. The evaluation of the 'CHEO best practice' rule did not consider those who received asthma education or attended a respirology clinic visit as a positive outcome otherwise that would positively bias the results.

In this study we also reported the AUC to enable comparison with previous studies, although the AUC does not consider the optimal cutoff for decision making, especially in a real-world setting.

## 2.7  ML Model Explainability

SHAP (Shapley Additive exPlanations) [76] values for the models were approximated using Kernel SHAP [77], where the contribution of each feature for asthma ED visit or hospital admission prediction are visualized. Model explainability was only applied to the most accurate prediction model.

## 3.  Results

### 3.1  Descriptive Summary

There were a total of 2716 patients in the training cohort and 1237 in the validation cohort. The distribution of patient characteristics was similar between cohorts with the majority being preschoolers (mean age 4.4-4.5 years, median 3 years), 34.3-36.5% being female, and the majority (72.4% - 81.8%) residing in Ontario (see Table 1).

|  | **Training cohort (n = 2716)** | **Validation cohort (n = 1237)** |
|---|---|---|
| Sex | 34.3% female<br>0.2% missing | 36.5% female<br>0.2% missing |
| Province of Residence | 72.4% Ontario<br>26.4% Quebec<br>1% other<br>0.2% missing | 81.8% Ontario<br>17.4% Quebec<br>0.6% other<br>0.2% missing |
| Mean Age in years (median) | 4.5 (3)<br>0.2% missing | 4.4 (3)<br>0.2% missing |
| Has a primary care provider | 94.5% | 96.4% |

**Table 1:** Baseline demographics of the training cohort and validation cohort.

Outcome distributions were also similar between cohorts, ranging from 29.3% - 28.5% repeat asthma ED visits in the training and validation cohorts respectively, as shown in Table 2.

|  | Patients with ED Visit outcome N (%) | Patients with admission outcome N (%) |
|---|---|---|
| Training (Feb 2017 – Feb 2019) N = 2716 | 795 (29.3) | 323 (11.9) |
| Validation (Jul 2022 – Apr 2023) N = 1237 | 352 (28.5) | 220 (17.8) |

**Table 2:** Asthma-related Emergency Department visit or hospital admission outcome distributions among training and validation cohorts. ED = emergency department.

### 3.2 Baseline Performance

We compared models against 3 different baselines as defined above. Table 3 shows the performance of the current "CHEO best practice" rule for alerting ED clinicians about which patients to refer to the CHEO asthma program. The performance of the other two baselines is shown in Supplementary materials.

| Current CHEO Best Practice Alert | Training | | Validation | |
|---|---|---|---|---|
|  | ED Visit | Admission | ED Visit | Admission |
| Precision | 0.467 | 0.248 | 0.528 | 0.368 |
| Recall | 0.190 | 0.248 | 0.244 | 0.273 |
| F1 | 0.270 | 0.248 | 0.334 | 0.313 |

**Table 3:** Performance of the current 'CHEO best practice' rule for identifying patients for referral to the CHEO asthma program based on an assumed higher risk of repeat asthma-related hospitalization. CHEO = Children's Hospital of Eastern Ontario, ED = emergency department, F1 = a score that optimizes both precision and recall to evaluate performance of an algorithm or model.

### 3.3 ML Model Performance

We first present the results of internal validation only on the training dataset using nested cross validation with 5-folds in the inner and outer loops. The AUC values for the various models on the training dataset ranged from 0.530 to 0.666 for predicting ED visits and 0.5 to 0.74 for predicting admissions, with lower AUC's observed for the LLMs as shown in Table 4. The best performance is for the LGBM model.

| ML Model class | ML model | AUC for ED visit | AUC for hospital admission |
|---|---|---|---|
| Boosted tree-based models | LGBM | 0.666 | 0.742 |
| | XGB | 0.657 | 0.737 |
| Large language models | Distilgpt2 | 0.587 | 0.645 |
| | Llama1b | 0.530 | 0.500 |
| | Llama8b | 0.582 | 0.650 |

**Table 4:** Assessing area under the curve for various ML models using nested cross validation on the training dataset. Note that these results do not include model calibration as beta calibration does not have an impact on AUC. ML = machine learning, ED = emergency department, AUC = area under the curve.

Using the models derived on the training dataset, F1 scores for predicting ED visits ranged from 0.292 to 0.503, and from 0.140 to 0.342 for predicting admissions when applied to the validation dataset. In all cases, F1 scores were highest for the LGBM models, and lowest for the LLMs as shown in Table 5.

|  | ED Visits | | | | | Admissions | | | | |
| --- | --- | --- | --- | --- | --- | --- | --- | --- | --- | --- |
|  | threshold | AUC | Precision | Recall | F1 | threshold | AUC | Precision | Recall | F1 |
| LGBM | 0.270 | 0.702 | 0.387 | 0.719 | 0.503 | 0.280 | 0.673 | 0.472 | 0.268 | 0.342 |
| XGB | 0.150 | 0.708 | 0.357 | 0.818 | 0.497 | 0.050 | 0.661 | 0.521 | 0.227 | 0.316 |
| Distilgpt2 | 0.050 | 0.699 | 0.638 | 0.170 | 0.269 | 0.050 | 0.660 | 0.574 | 0.123 | 0.202 |
| Llama1b | 0.060 | 0.609 | 0.370 | 0.528 | 0.435 | 0.180 | 0.548 | 0.373 | 0.086 | 0.140 |
| Llama8b | 0.050 | 0.582 | 0.452 | 0.216 | 0.292 | 0.100 | 0.633 | 0.456 | 0.214 | 0.291 |

**Table 5**: Evaluation of training dataset-derived models when applied to the validation dataset. ED = emergency department, threshold = the probability cutoff for defining a positive or negative outcome. AUC = area under the curve. F1 = reflects both the precision and recall. Boosted tree-based models included LGBM, XGB; large language models included Distilgpt2, Llama1b, Llama8b.

## 3.4 Model Explainability

The features most predictive for repeat asthma ED visits based on SHAP plots (see Figure 3) for the LGBM model included previous asthma ED visit, CTAS, medical complexity, food allergy, ED visit for other respiratory diagnoses, and age. Similarly, for predicting future asthma hospitalization (Figure 4) for the LGBM model, the top features included medical complexity, previous asthma ED visit, average wait time, PRAM score during the index visit and food allergy.

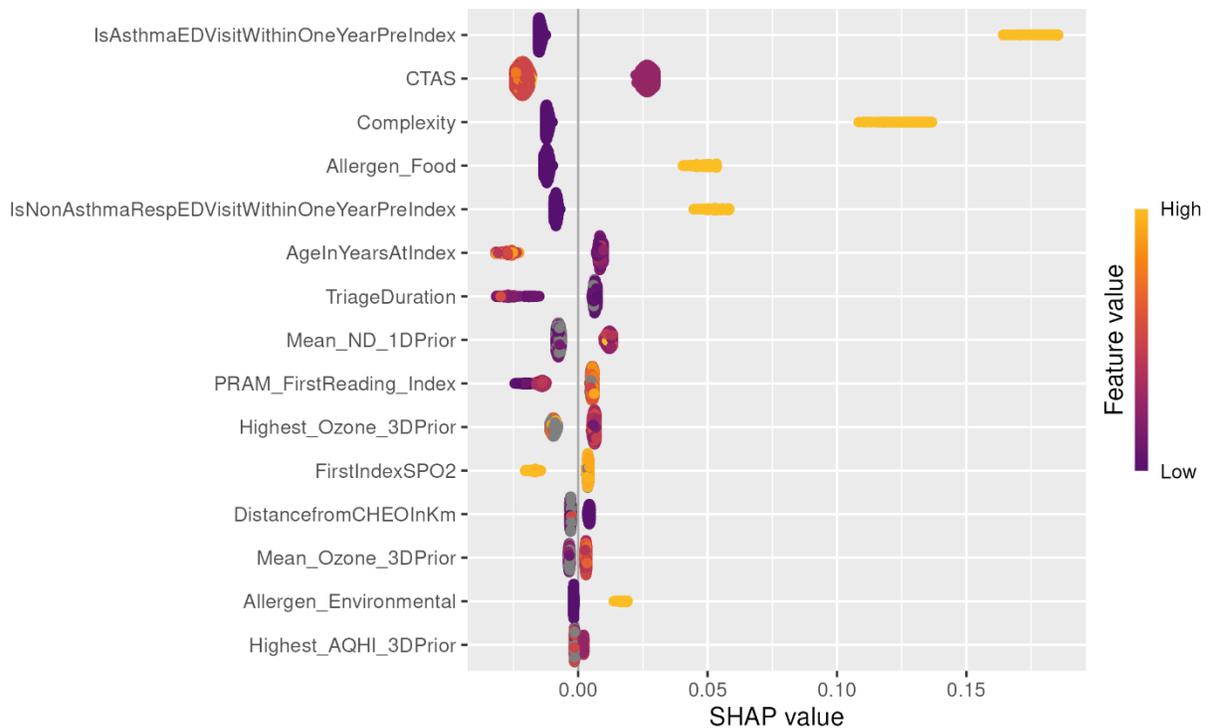

**Figure 3:** SHAP values for predicting repeat asthma ED visits based on the training dataset and the LGBM model. SHAP = Shapley Additive exPlanations, ED = emergency department, CTAS = Canadian triage acuity scale, ND = nitrogen dioxide, 1D = 1 day, PRAM = pediatric respiratory assessment measure, 3D = 3 days, SPO2 = oxygen saturation, AQHI = air quality health index.

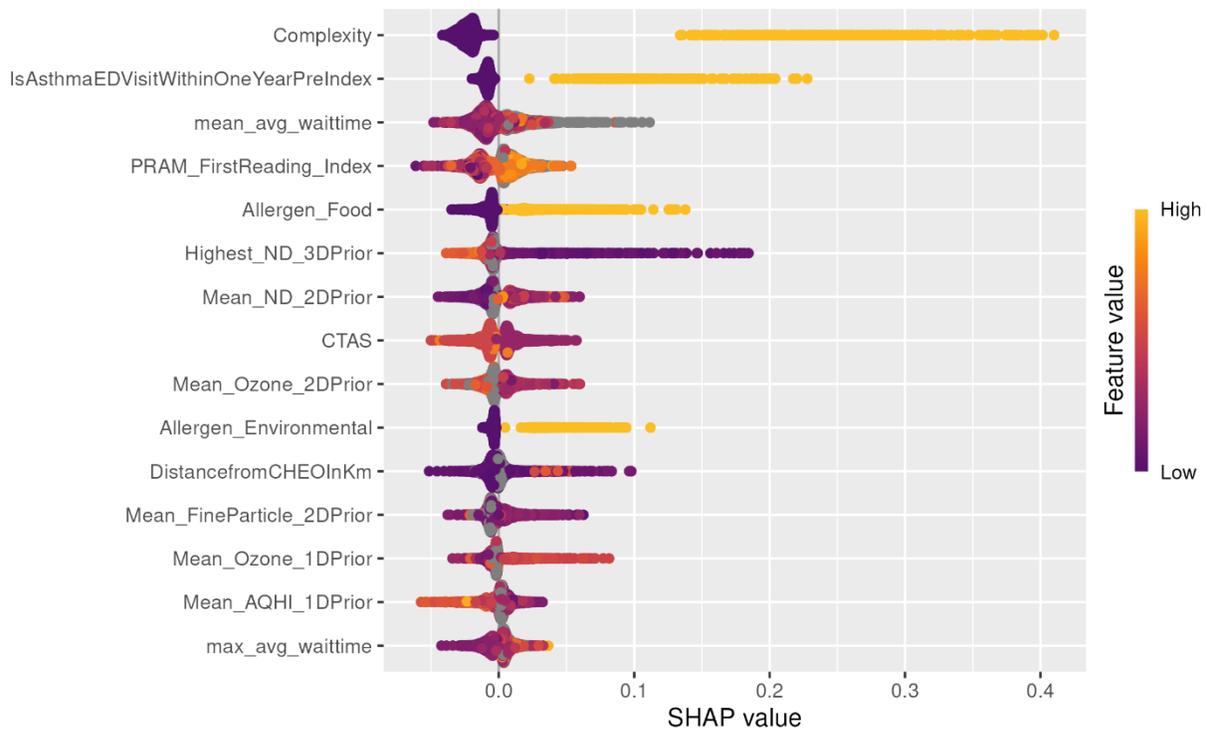

**Figure 4:** SHAP values for predicting future asthma hospitalization based on the training dataset and the LGBM model. SHAP = Shapley Additive exPlanations, ED = emergency department, avg = average, PRAM = pediatric respiratory assessment measure, ND = nitrogen dioxide, 3D = 3 days, 2D = 2 days, CTAS = Canadian triage acuity scale, CHEO = Children's Hospital of Eastern Ontario, Km = kilometres, 1D = 1 day

### 3.5 Final AIRE-KIDS Models

Based on the feature importance from the SHAP analysis, we chose a smaller set of features for both outcomes based on the variable importance results, And retrained the LGBM models. These more parsimonious models would be easier to deploy in practice and would have similar performance as the models with all of the features included.

For the ED visit outcome, the selected features in the final AIRE-KIDS$_{ED}$ model based on the highest ranked predictors in Figure 3 were: prior asthma ED visit, CTAS, complexity, food allergy, prior ED visits for non-asthma respiratory diagnoses, and age. Using these features only and maximizing the AUC value on the training dataset with nested cross-validation we obtained an AUC of 0.712, and F1 of 0.51 at a cutoff threshold of 0.250.

For the future asthma admission outcome, the final AIRE-KIDS$_{HOSP}$ model features based on the top ranked predictors in Figure 4 were: complexity, prior asthma ED visit, average wait time in the ED, PRAM at triage and food allergy. Using these features only and maximizing the AUC value on the training dataset with nested cross-validation we obtained an AUC of 0.65, and F1 score of 0.375 at a cut threshold of 0.170 (see Table 6).

|  | Training Dataset | Validation Dataset | | | | |
|---|---|---|---|---|---|---|
|  | AUC | threshold | AUC | Precision | Recall | F1 |
| ED Visits AIRE-KIDS$_{ED}$ | 0.679 | 0.250 | 0.712 | 0.371 | 0.815 | 0.510 |
| Admission AIRE-KIDS$_{HOSP}$ | 0.739 | 0.170 | 0.65 | 0.317 | 0.459 | 0.375 |

**Table 6:** Final performance of both AIRE-KIDS models with reduced features. AUC = area under the curve, F1 = reflects both precision and recall, ED = emergency department, HOSP = hospital admission.

## 4. Discussion
### 4.1 Summary
This study describes the development of the AIRE-KIDS ML models, designed to predict future severe asthma exacerbations. These were defined as: repeat asthma emergency department or hospital admissions among children presenting to a tertiary care pediatric ED. The LGBM model yielded the best performance on a large validation dataset demonstrating good discrimination with an AUC of 0.68 and 0.74 for predicting repeat asthma ED visits or future hospital admission, respectively. These models additionally had greatly improved performance compared to the current ED-based best practice alert in use at the hospital with F1 scores of 0.51 vs 0.334 and 0.375 vs 0.313 respectively. The best naïve decision rule that predicts all patients having high risk has an F1 score of 0.442 and 0.302 respectively. Therefore, the deployment of the LGBM model would result in a meaningful improvement in the decision-making process.

AIRE-KIDS also outperformed all tested LLM-based prediction approaches. While LLMs are gaining popularity as they offer flexible and broad applications and can be easier to deploy, they lack the precision required for targeted clinical decision-making. The LGBM model used here offers a task-specific, interpretable alternative, with greater transparency and trustworthiness critical for clinical adoption.

This improved performance represents a major opportunity to more effectively triage children who would most benefit from comprehensive asthma care to reduce the volume of repeat acute healthcare visits at our centre.

While the AUC is a traditional benchmark for ML discriminative performance, this approach is not satisfactory in our clinical decision-making contexts where system constraints and resource limitations must be considered. For example, the CHEO asthma program is limited in its annual capacity to accommodate new referrals, and not every child predicted to have a repeat asthma ED visit or hospitalization could be included in the program. The prediction objective is to focus on positive cases and maximize precision to ensure that referrals are indeed at high risk of asthma exacerbation and that limited resources are used most efficiently. However, to avoid training trivial models we must simultaneously maximize recall.

Another pitfall of using the AUC alone to judge model performance is that it does not properly reflect imbalanced datasets where the positive outcome (e.g. hospital admission) is relatively rare. In this context, for AIRE-KIDS, a balance between both recall and precision was preferred, and the F1 score was prioritized for choosing the best performing ML models. As such the AIRE-KIDS$_{ED}$ model could be considered the optimal model for clinical implementation at our centre given its higher F1 score. However, as future hospital admission is a more important outcome that we would not want to miss, then the AIRE-KIDS$_{HOSP}$ model could also be implemented.

The most predictive features for predicting repeat ED visits in the AIRE-KIDS$_{ED}$ model included previous asthma-related ED visits, medical complexity, triage acuity (using CTAS), food allergy, prior non-asthma ED visits and age. The PRAM score at triage and average ED wait time were additional key features in the AIRE-KIDS$_{HOSP}$ for predicting future hospitalization. Although environmental and social variables (e.g., air quality, marginalization index) were considered, they did not significantly contribute to model performance and were excluded from the final models to improve parsimony and clinical utility.

There is a large literature on predicting asthma exacerbations in children [78–80], with previous studies ranging from descriptions of patient-facing algorithms that require direct input of symptom data (e.g. questionnaires [11,12]) to predict short term deterioration or immediate hospital admission from the ED [16], to studies involving large health administrative data [7] to predict exacerbation in a longer timeframe. While these latter studies are similar to ours and could be used to identify children at risk for future exacerbations to enable optimization of preventative asthma management, many rely on predictors that may not be immediately available at the ED point of care. For example, Wang *et al.* looked at claims data for 28,378 pediatric patients and developed ML models with higher AUC's (0.84) than AIRE-KIDS but lower F1 scores (0.2) for predicting repeat ED visit in 3 months [14]. These models additionally relied on administrative data which typically has a lag-time before being available to clinicians, making real-time deployment challenging. In contrast, AIRE-KIDS achieves a more balanced performance using real-time data at the point of care, increasing its feasibility for clinical deployment directly from the ED.

Our study methods and findings align with those by Das *et al*.[3], and Gorham *et al.* [13] who also developed ML models using EHR data from US centres to predict asthma-related future ED visits over a 1-year window. While these models showed comparable performance (AUC 0.76 – 0.86) and identified certain key features that are similar to AIRE-KIDS (e.g. previous ED use, age), other features such as insurance status were not applicable to our models. These differences in important feature highlight contextual differences between countries and healthcare systems and demonstrate the challenges in broad generalizability of clinical ML models and the importance of locally trained and validated models for most effective implementation.

## 4.2 Limitations

One limitation to the AIRE-KIDS models is whether they would be generalizable to other centres, and this would require additional validation to ascertain. Likely additional training or calibration would be required to reflect local case mix and practice variation, even within a similar healthcare system [13]. However, by prioritizing transparency, and parsimony with fewer variables to allow for easy clinical deployment and integration into EMRs with limited compute available at the point-of-care, AIRE-KIDS offers a practical solution for real-world settings that could likely be scaled to other centres with minimal retraining.

An additional limitation of our study includes that we were restricted to using local healthcare visit data and may have missed some outcomes when children visited other centres for asthma exacerbations. However, as more than 94% of all children living within a 50km radius attend CHEO for their acute care, the magnitude of these missed outcomes is likely small, and would not have significantly changed results.

Certain variables were missed from inclusion as they are difficult to obtain from EMR data, including adherence to asthma controller medications or exposure to cigarette smoke in the home; nonetheless even without these variables the performance of the AIRE-KIDS models offer a significant improvement over current practice in our ED and would be considered adequate for clinical adoption.

Including patients who received asthma education or attended a respirology clinic visit in the outcome variables introduces error. As seen in the results, the existing rule for these referrals is not very accurate in detecting high risk patients. Therefore, our results are arguably better than presented here because of this. Notwithstanding this limitation, the prediction models we developed demonstrated nontrivial improvements over the current baseline and naïve prediction.

## 5. Conclusions

AIRE-KIDS$_{ED}$ and AIRE-KIDS$_{HOSP}$ are new ML models for accurately predicting repeat asthma related ED visits or future asthma-related hospitalizations in children, and are superior to current decision-making processes for triaging referrals for comprehensive preventative care from the ED. LGBM remains a superior ML approach for modeling this type of clinical prediction, outperforming newer LLMs, despite their popularity in other applications, though ongoing research in this area is needed as LLM's continue to evolve.

The next steps to realize the clinical utility of the AIRE-KIDS models would involve local deployment and prospective evaluation at our centre in the form of a clinical decision support system and subsequent validation at other pediatric centres. Successful and broad implementation of AIRE-KIDS could redefine how we triage and manage pediatric asthma in the acute care setting, supporting more equitable access to comprehensive or specialist care while reducing the morbidity associated with repeat severe exacerbations that result in ED visits and hospitalizations and their burden on the healthcare system, children and families.

## Ethics

This study was approved by the Research Ethics Board of the Children's Hospital of Eastern Ontario Research Institute under protocol number CHEOREB# 22/11X on 14$^{th}$ February 2022.

## Consent To Participate

This study involved the secondary analysis of de-identified data for internal analysis by the hospital research institute only, and therefore did not require participant consent. The consent requirement was waived by the research ethics board.

## Clinical Trial Number

Not Applicable.

## Acknowledgements

The authors would like to acknowledge Ivan Terekhov and Jon Seymour for Epic EMR data extraction for this study, and in-kind support provided by the CHEO Research Institute Informatics Core.

## Funding Statement

Funding for this study was provided by the Children's Hospital Academic Medical Organization (CHAMO), Canada Research Chairs program through the Canadian Institutes of Health Research, a Discovery Grant RGPIN-2022-04811 from the Natural Sciences and Engineering Research Council of Canada, and Mitacs.

## Competing Interests Statement

The authors have no competing interests to report.

## Author Contributions Statement

Conceptualization and design: HLO, NM, KEE, DR, JC, RZ, and AP; analysis: HLO, NM, KEE, DR, MHD, and JC; writing and review: HLO, NM, DR, KEE, MHD, JC, RZ, and AP.

# APPENDIX

# AI for pRedicting Exacerbations in KIDs with aSthma (AIRE-KIDS)

# 1. Appendix A: Literature Summary

This appendix includes a tabular summary of previous related work.

## 1.1 Outcome Definition

The following table Table S1 summarized the outcomes that were used in previous related studies.

| Reference | Oral prescription | ED Visit | Hospitalization | Exacerbation | Death |
|---|---|---|---|---|---|
| [1] | Non severe (28 days) | Severe | severe | | |
| [2] | Moderate (7 days) | Severe (15 days) | Severe (15 days) | | |
| [3] | | 1 year | 1 year | | |
| [4] | | 1 year | 1 year | | |
| [5,6] | | 1 year | 1 year | | |
| [7] | | 1 year | | | |
| [8] | | 3 months | | | |
| [9] | | | 180 days | | |
| [10] | | | ✓ | | |
| [11] | | | ✓ | | |
| [12] | | | ✓ | | |
| [13] | | | 1 year | | |
| [14] | | | ✓ | | |
| [15] | 1 year | 1 year | 1 year | | |
| [16] | 1 year | 1 year | 1 year | | |
| [17] | | 30 days, 90 days, 180 days | 30 days, 90 days, 180 days | | |
| [18] | 90 days | 90 days | 90 days | | |
| [19] | 1 week, 4 weeks, 26 weeks, 52 weeks | 1 week, 4 weeks, 26 weeks, 52 weeks | 1 week, 4 weeks, 26 weeks, 52 weeks | | |
| [20] | | | 1 year | | 1 year |
| [21] | 30 days | 30 days | 30 days | | 30 days |
| [22] | | | | ✓ | |

**Table S1:** Outcome Definitions and Prediction Window Length

## 1.2    Countries of Study

The following Table S2 are the countries where previous studies were conducted.

| US | [3–10,13–18,21,23,23] |
|---|---|
| Canada | [24–28] |
| Sweden | [2,29] |
| Scotland | [19,20] |
| Croatia | [30] |
| Greece | [31–35] |
| Japan | [36,37] |
| New Zealand | [38–40] |

**Table S2:** Related Studies across Different Countries.

## 1.3    Performance Measures

The summary on Table S3 gives the performance measures that were used to evaluate previous prediction models.

|  | AUC | Accuracy | Sensitivity | Specificity | PPV | NPV |
|---|---|---|---|---|---|---|
| Tong et al [4] | ✓ | ✓ | ✓ | ✓ | ✓ | ✓ |
| Luo et al [5,6] | ✓ | ✓ | ✓ | ✓ |  |  |
| Sato et al [37] | ✓ | ✓ |  |  |  |  |
| Schatz et al [41,42] | ✓ | ✓ | ✓ | ✓ | ✓ | ✓ |
| Lieu et al [3,43] | ✓ | ✓ |  |  |  |  |
| Tibble et al [19] | ✓ | ✓ | ✓ | ✓ | ✓ | ✓ |
| Noble et al [20] | ✓ |  | ✓ | ✓ |  |  |
| Bose et al [23] |  | ✓ | ✓ |  | ✓ | ✓ |
| Owora et al [28] | ✓ | ✓ | ✓ |  |  |  |
| Hurst et al [17] | ✓ |  | ✓ |  | ✓ |  |
| Zein et al [1] | ✓ | ✓ | ✓ | ✓ | ✓ | ✓ |
| Patel et al [11] | ✓ |  |  |  |  |  |
| Wang et al [8] | ✓ |  |  |  |  |  |
| Hogan et al [9] | ✓ |  |  |  |  |  |
| Sills et al [10] | ✓ | | | | | |
| Goto et al [12] | ✓ | | ✓ | ✓ | ✓ | ✓ |
| Barak-Corren et al [14] | ✓ | ✓ | ✓ | ✓ | ✓ | ✓ |
| Xiang et al [44] | ✓ | | | | | |
| Loymans et al [45] | ✓ | | | | | |
| Kothalawala et al [46] | ✓ | | | | | |
| Overgaard et al [15] | ✓ | | | | | |
| Hatoun et al [16] | ✓ | | | | | |
| Cobian et al [18] | ✓ | | | | | |
| Halner et al [21] | ✓ | | | | | |
| Jiao et al [27] | ✓ | | | | | |
| de Hond et al [40] | ✓ | | | | | |

PPV: positive predictive value, also known as precision

NPV: negative predictive value

**Table S3:** Evaluation metrics used in Related Studies.

## 1.4 Predictors Used

The following Table S4 summarizes the predictors that were used in previously fitted models.

| | [14] | [23] | [13] | [7] | [16] | [9] | [17] | [3] | [2] | [5] | [11] | [10] | [19] | [4] | [8] | [1] | [20] | [12] | [30] | [21] |
|---|---|---|---|---|---|---|---|---|---|---|---|---|---|---|---|---|---|---|---|---|
| Age | ✓ | ✓ | ✓ | ✓ | ✓ | ✓ | ✓ | ✓ | ✓ | ✓ | ✓ | ✓ | ✓ | ✓ | ✓ | ✓ | ✓ | ✓ | | |
| Previous visit/hospitalization | ✓ | ✓ | | ✓ | | | ✓ | | | | | ✓ | ✓ | ✓ | ✓ | | ✓ | | | ✓ |
| Sex | ✓ | ✓ | | ✓ | ✓ | ✓ | ✓ | ✓ | | ✓ | ✓ | | | | | | ✓ | ✓ | | ✓ |
| BMI | ✓ | | | | | | | ✓ | | | | ✓ | | | ✓ | ✓ | | | | |
| Medication history | ✓ | ✓ | ✓ | ✓ | ✓ | | ✓ | ✓ | ✓ | ✓ | | ✓ | ✓ | ✓ | ✓ | ✓ | ✓ | | | ✓ |
| Allergy | | ✓ | ✓ | | | | ✓ | | | ✓ | | | ✓ | | | | | | ✓ | |
| Comorbidity | | | | | | | ✓ | | ✓ | | | ✓ | ✓ | | | | ✓ | ✓ | | |
| Environmental features | | | | | | | ✓ | | | | | ✓ | | | | | | | | |
| Seasonal features | | | | | | ✓ | | | | ✓ | | | | | | | | | | |
| Socioeconomic features | | ✓ | ✓ | | | ✓ | ✓ | | ✓ | ✓ | ✓ | | | | | | | | | |
| Lab tests features | ✓ | | | | | | | | | ✓ | | | ✓ | | ✓ | ✓ | | | ✓ | |
| Distance from hospital features | ✓ | | | | | | ✓ | | | | | | | | | | | | | |
| Arrival mode | ✓ | | | | | | | | | | | ✓ | | | | | | ✓ | | |
| Hospital operational features | | | | | | | | | | | | ✓ | | | | | | | | |

Hospital operational features refer to ED treatment factors such as time to triage, time to first medication, variables reflecting timing and medication as well as ED crowding level.

**Table S4:** Predictors Used in Related Studies

## 1.5 Modeling Techniques Used

The following Table S5 is a summary of the modeling techniques used to predict the outcomes noted in Table S1.

| | |
|---|---|
| Logistic regression | [1,7,8,10–18,20,23] |
| Ensemble learning (LGBM) | [1,2,11,12] |
| Ensemble learning (XGBoost) | [2,4–6,17,23] |
| Ensemble learning (RF) | [1,2,10–12,12,15,17–19,21,23,30] |
| Ensemble learning (Adaboost) | [30] |
| Tree based (CART) | [3] |
| Tree based (decision tree) | [11] |
| SVM | [15,19] |
| KNN | [23] |
| Naïve bayes classifier | [14,15,19,23] |
| ANN | [8,9,12,15] |
| DL (RNN) | [2,44,47] |
| DL (LSTM) | [18] |

LGBM: Light Gradient-Boosting Machine
XGBoost: eXtreme Gradient Boosting
RF: Random Forest
Adaboost: Adaptive Boosting
CART: Classification and Regression Tree
ANN: Artificial Neural Network
SVM: Support Vector Machine
KNN: K Nearest Neighbor
RNN: Recurrent Neural Network
LSTM: Long Short-Term Memory

**Table S5:** Types of Models in Related Studies.

## 2. Appendix B: List of Features Included in Our Models

### 2.1 Boolean Features

The results in Table S6 provide summary information on the Boolean values in the features considered.

| Boolean Feature name | Definition | Pre-COVID Boolean features (True %) | Post-COVID Boolean features (True %) |
|---|---|---|---|
| **IsAdmissionWithinOneYearPreIndex** | Patient index encounter that has asthma diagnosis and is admitted within one year pre index, excluded patients that are admitted the same day as index encounter | 6.00 | 4.37 |
| **IsAsthmaEDVisitWithinOneYearPreIndex** | Patient with index encounter that has asthma diagnosis and has ED visit within one year pre index, excluded patients that are admitted the same day as index encounter | 7.62 | 9.94 |
| **IsNonAsthmaRespAdmissionWithinOneYearPreIndex** | Patient with index encounter that has asthma diagnosis and has non asthma resp admission within one year pre index, excluded patients that are admitted the same day as index encounter | 2.72 | 2.51 |
| **IsNonAsthmaRespEDVisitWithinOneYearPreIndex** | Patient with index encounter that has asthma diagnosis and has non asthma resp ED visit within one year pre index, excluded patients that are admitted the same day as index encounter | 13.77 | 13.42 |
| **HasPCP** | HasPCP (primary care physician) | 94.51 | 96.44 |
| **Complexity** | Presence of BMI from height and weight measurement in the patient's record during index visit | 8.58 | 9.94 |
| **PatientComorbiditiesPreIndex** | Patient with comorbidities before index encounter | 5.49 | 2.75 |
| **Is_Ventilated_Intubated_At** | Prior to and up until index, has the patient been intubated or ventilated (IntervDate == IndexDate) | 0.07 | 0.00 |
| **Allergen_Food** | Presence of patient's food allergy | 19.88 | 19.48 |
| **Allergen_Environmental** | Presence of patient's environmental allergy | 9.17 | 8.33 |

Missing values were indicated as False.

**Table S6:** Boolean features included in the full models.

## 2.2 Categorical Features

The categorical features included in our model are summarized in Table S7.

| Feature name | Definition | Percentage per categories | | missing% | |
| --- | --- | --- | --- | --- | --- |
| | | Pre-COVID | Post-COVID | Pre-COVID | Post-COVID |
| ArrivalMethod | The method of patient's arrival to the ED ['Car' 'Land Ambulance' 'Taxi' 'Walk' 'Unknown' 'Air Ambulance' 'Air & Ground Ambulance' 'Bus' '*Unspecified'] | 95.29% Car, 4.01% Land Ambulance, 0.26% Taxi, 0.18% Walk, 0.07% Unknown, 0.04% Air Ambulance, 0.04% Air & Ground Ambulance, 0.04% Bus, 0.07% *Unspecified | 94.02% Ambulatory, 5.50% Land Ambulance, 0.08% Air Ambulance, 0.32% Inter-facility Transfer, 0.08% Air & Ground Ambulance | 0 | 0 |
| Index_Month | Month of index visit date | 6.70% Jan, 12.19% Feb, 9.32% Mar, 10.13% Apr, 9.13% May, 0% Jun, 4.97% Jul, 4.75% Aug, 10.94% Sept, 9.02% Oct, 8.06% Nov, 7.22% Dec | 8.00% Jan, 8.00% Feb, 6.87% Mar, 6.87% April, 7.58% May, 0% Jun, 7.19% Jul, 10.67% Aug, 23.20% Sept, 12.85% Oct, 9.22% Nov, 7.11% Dec | 0 | 0 |
| Index_Day | Day of the week for the index visit date | 16.94% Mon, 13.99% Tue, 12.33% Weds, 12.96% Thurs, 11.27% Fri, 15.24% Sat, 17.27% Sun | 14.39% Mon, 13.90% Tues, 14.23% Weds, 14.55% Thurs, 13.66% Fri, 14.71% Sat, 14.55% Sun | 0 | 0 |
| CTAS | From acuity level ['2 - EMERGENT' '3 - URGENT' '5 - NON-URGENT' '4 - SEMI-URGENT' '1 - RESUS'] | 0.92% level 1, 44.70% level 2, 41.49% level 3, 9.57% level 4, | 2.34% level 1, 50.77% level 2, 39.94% level 3, 4.37% level 4, | 0 | 0.08 |

|  |  | 3.31% level 5 | 2.51% level 5 |  |  |
|---|---|---|---|---|---|
| **Sex** | Sex of patients | 65.50% Male, 34.32% Female | 63.22% Male, 36.54% Female | 0.18 | 0.24 |
| **Province** | Province of patient's address, derived from postal code | 72.42% ON, 26.36% QC, 1.03% Others | 81.81% ON, 17.38% QC, 0.57% Others | 0.18 | 0.24 |
| **households_dwellings** | Marginalization on household dwelling<br>Quintiles - Households and dwellings (1=low household and dwelling-related marginalization, 5=high household and dwelling-related marginalization) | 0.4% level 1, 4.16% level 2, 0.81% level 3, 1.18% level 4, 65.72% level 5 | 0.65% level 1, 4.85% level 2, 1.05% level 3, 1.62% level 4, 72.51% level 5 | 27.72 | 19.32 |
| **material_resources** | Marginalization on material resources<br>Quintiles - Material resources (1=low material resources-related marginalization, 5=high material resources-related marginalization) | 4.64% level 1, 65.21% level 2, 0.77% level 3, 0.55% level 4, 1.10% level 5 | 5.09% level 1, 70.98% level 2, 1.37% level 3, 0.97% level 4, 2.26% level 5 | 27.72 | 19.32 |
| **age_labourforce** | Marginalization on labour force Quintiles - Age and labour force (1=low age and Comordibity_K21_9 labour force-related marginalization, 5=high age and labour force-related marginalization) | 66.02% level 1, 3.94% level 2, 0.88% level 3, 0.81% level 4, 0.63% level 5 | 71.87% level 1, 5.25% level 2, 0.65% level 3, 2.43% level 4, 0.49% level 5 | 27.72 | 19.32 |
| **racial_newcomer_population** | Marginalization on racialized population<br>Quintiles - Racialized and newcomer populations (1=low racialized and newcomer population-related | 0.22% level 1, 0.55% level 2, 1.84% level 3, 3.68% level 4, 65.98% level 5 | 0.08% level 1, 0.24% level 2, 3.4% level 3, 4.12% level 4, 72.84% level 5 | 27.72 | 19.32 |

| | marginalization, 5=high racialized and newcomer population-related marginalization) | | | | |
|---|---|---|---|---|---|

Table S7: Categorical features included in the full models.

## 2.3 Continuous Features

The continuous features included in our models are summarized in Table S8.

| Feature name | Definition | Pre-COVID | | | Post-COVID | | |
|---|---|---|---|---|---|---|---|
| | | missing% | mean | median | missing% | mean | median |
| **TriageDuration** | Time taken from the start till the end of triage in minutes | 40.21 | 7.01 | 4.87 | 1.37 | 8.63 | 6.69 |
| **AgeInYearsAtIndex** | Age during index encounter | 0.18 | 4.49 | 3.00 | 0.24 | 4.41 | 3.00 |
| **DistancefromCHEOInKm** | Straight line distance from postal code of home to hospital in km | 27.69 | 20.19 | 11.86 | 19.16 | 22.37 | 13.62 |
| **Arrival2DispositonDuration** | Duration of index visit starting from the arrival instant until the disposition instant | 38.70 | 4.75 | 3.84 | 0 | 6.50 | 4.52 |
| **Max_num_patients** | Max number of patients among the hourly interval throughout the patient's encounter visit | 38.70 | 26.38 | 26.00 | 0 | 35.56 | 35.00 |
| **Mean_num_patients** | Mean number of patients among the hourly interval throughout the patient's encounter visit | 38.70 | 20.62 | 19.50 | 0 | 28.86 | 28.00 |
| **Max_avg_waittime** | Max average waittime among the hourly interval throughout the patient's encounter visit | 38.70 | 109.72 | 101.00 | 0 | 199.76 | 165.00 |
| **Mean_avg_waittime** | Mean average waittime among the hourly interval throughout the patient's encounter visit | 38.70 | 83.90 | 79.14 | 0 | 143.40 | 122.17 |
| **MaxEosinophils** | Highest lab value for 'Eosinophils', before or until index date | 86.19 | 0.31 | 0.20 | 96.20 | 0.76 | 0.20 |

| Variable | Description | | | | | |
|---|---|---|---|---|---|---|
| MaxIgE | Highest lab value for 'IgE, before or until index date | 99.23 | 1274.67 | 104.00 | 99.68 | 1054.00 | 1005.50 |
| PRAM_FirstReading_Index | first PRAM score read during index visit (0-3: mild, 4-7: moderate, 8-12: severe) | 44.07 | 5.89 | 6.00 | 3.80 | 6.51 | 8.00 |
| PRAM_FirstReading_greater 3h | first PRAM score read during index visit after the first 3h (0-3: mild, 4-7: moderate, 8-12: severe) | 61.52 | 2.77 | 3.00 | 24.74 | 2.88 | 3.00 |
| FirstIndexSPO2 | First reading of SPO2 during patient's index visit | 39.18 | 95.74 | v96.00 | 14.23 | 94.56 | 95.00 |
| Highest_ND_1DPrior | Highest NitrogenDioxide 1 day prior index (from 24h up until 1h before index visit) | 27.76 | 14.85 | 12.00 | 43.57 | 10.13 | 7.80 |
| Highest_ND_2DPrior | Highest NitrogenDioxide 2 day prior index (from 48h up until 1h before index visit) | 27.76 | 18.71 | -15.30 | 43.57 | 12.45 | 10.70 |
| Highest_ND_3DPrior | Highest NitrogenDioxide 3 day prior index (from 72h up until 1h before index visit) | 27.76 | 21.10 | 18.20 | 43.57 | 13.99 | 11.60 |
| Highest_Ozone_1DPrior | Highest Ozone 1 day prior index (from 24h up until 1h before index visit) | 27.76 | 36.40 | 36.00 | 43.57 | 31.81 | 31.00 |
| Highest_Ozone_2DPrior | Highest Ozone 2 day prior index (from 48h up until 1h before index visit) | 27.76 | 39.41 | 39.00 | 43.57 | 35.29 | 34.00 |
| Highest_Ozone_3DPrior | Highest Ozone 3 day prior index (from 72h up until 1h before index visit) | 27.76 | 41.30 | 40.00 | 43.57 | 37.64 | 35.00 |

| Variable | Description | N | Mean | Median | N | Mean | Median |
|---|---|---|---|---|---|---|---|
| Highest_FineParticle_1DPrior | Highest FineParticulateMatter 1 day prior index (from 24h up until 1h before index visit) | 27.76 | 9.62 | 8.00 | 43.57 | 8.73 | 8.00 |
| Highest_FineParticle_2DPrior | Highest FineParticulateMatter 2 day prior index (from 48h up until 1h before index visit) | 27.76 | 11.81 | 10.00 | 43.57 | 11.54 | 10.00 |
| Highest_FineParticle_3DPrior | Highest FineParticulateMatter 3 day prior index (from 72h up until 1h before index visit) | 27.76 | 13.23 | 12.00 | 43.57 | 12.99 | 12.00 |
| Highest_AQHI_1DPrior | Highest AQHI (AirQualityHealthIndex) 1 day prior index | 27.76 | 2.84 | 3.00 | 43.57 | 2.33 | 2.00 |
| Highest_AQHI_2DPrior | Highest AQHI (AirQualityHealthIndex) 2 day prior index | 27.76 | 3.12 | 3.00 | 43.57 | 2.59 | 2.00 |
| Highest_AQHI_3DPrior | Highest AQHI (AirQualityHealthIndex) 3 day prior index | 27.76 | 3.31 | 3.00 | 43.57 | 2.76 | 3.00 |
| Mean_ND_1DPrior | Mean of NitrogenDioxide 1 day prior index (from 24h up until 1h before index visit) | 27.76 | 7.00 | 5.48 | 43.57 | 4.82 | 3.62 |
| Mean_ND_2DPrior | Mean of NitrogenDioxide 2 day prior index (from 48h up until 1h before index visit) | 27.76 | 7.04 | 5.74 | 43.57 | 4.73 | 3.80 |
| Mean_ND_3DPrior | Mean of NitrogenDioxide 3 day prior index (from 72h up until 1h before index visit) | 27.76 | 7.03 | 5.88 | 43.57 | 4.73 | 3.75 |
| Mean_Ozone_1DPrior | Mean of Ozone 1 day prior index (from 24h up until 1h before index visit) | 27.76 | 26.39 | 26.38 | 43.57 | 21.10 | 20.78 |

| Variable | Description | | | | | | |
|---|---|---|---|---|---|---|---|
| Mean_Ozone_2DPrior | Mean of Ozone 2 day prior index (from 48h up until 1h before index visit) | 27.76 | 26.43 | 26.18 | 43.57 | 21.40 | 20.20 |
| Mean_Ozone_3DPrior | Mean of Ozone 3 day prior index (from 72h up until 1h before index visit) | 27.76 | 26.51 | 26.26 | 43.57 | 21.48 | 20.85 |
| Mean_FineParticle_1DPrior | Mean of FineParticulateMatter 1 day prior index (from 24h up until 1h before index visit) | 27.76 | 5.71 | 4.83 | 43.57 | 5.08 | 4.70 |
| Mean_FineParticle_2DPrior | Mean of FineParticulateMatter 2 day prior index (from 48h up until 1h before index visit) | 27.76 | 5.78 | 5.00 | 43.57 | 5.21 | 4.90 |
| Mean_FineParticle_3DPrior | Mean of FineParticulateMatter 3 day prior index (from 72h up until 1h before index visit) | 27.76141 | 5.78 | 5.06 | 43.57 | 5.24 | 4.87 |
| Mean_AQHI_1DPrior | Mean of AQHI (AirQualityHealthIndex) 1 day prior index | 27.76141 | 2.22 | 2.17 | 43.57 | 1.72 | 1.70 |
| Mean_AQHI_2DPrior | Mean of AQHI (AirQualityHealthIndex) 2 day prior index | 27.76141 | 2.23 | 2.23 | 43.57 | 1.74 | 1.70 |
| Mean_AQHI_3DPrior | Mean of AQHI (AirQualityHealthIndex) 3 day prior index | 27.76141 | 2.23 | 2.24 | 43.57 | 1.75 | 1.72 |
| OCS_preindex_total_count | Total count of OCS ordered and administrated before index visit | 0 | 0.09 | 0.00 | 0 | 0.13 | 0.00 |
| OCS_index_total_count | Total count of OCS ordered and administrated during index visit | 0 | 0.52 | 0.00 | 0 | 0.88 | 1.00 |
| OCS_ordered_not_administrated_atindex_count | Count of OCS ordered at index visit but not administrated | 0 | 0.455081 | 0.00 | 0 | 0.65 | 0.00 |
| OCS_ordered_not_administrated_preindex_count | Count of OCS ordered at pre visit but not administrated | 0 | 0.04676 | 0.00 | 0 | 0.06 | 0.00 |
| **OCS_ordered_admin_3h_count** | Count of OCS ordered on the first 3h of index visit and also administered on the first 3h | 0 | 0.426362 | 0.00 | 0 | 0.72 | 1.00 |
| **OCS_ordered_admin_greater3h_count** | Count of OCS ordered after the first 3h of index visit and also administered after the first 3h | 0 | 0.079161 | 0.00 | 0 | 0.14 | 0.00 |
| **OCS_ordered3h_admingreater3h_count** | Count of OCS ordered at the first 3h of index visit and but administered after the first 3h | 0 | 0.01215 | 0.00 | 0 | 0.02 | 0.00 |

AQHI [48] is a scale to measure air quality by Environment Canada, Health Canada and the Province of Ontario where 1-3 are considered low risk, 4-6 are considered moderate risk, 7-10 are considered high risk and above 10 is considered very high risk. Ozone [49] acts as a pollutant at the ground level and is the primary component of smog. NO2 [50] are highly reactive gases, entering the air from fuel combustion and therefore indicates air pollution from motor vehicles. PM2.5 [51] refers to the fine particulate matter that are less than 2.5 micrometers in diameter in the air and may be a mix of smoke, soot, liquid or solid particles.

Marginalization is defined as the process where certain individuals or groups experiencing barrier or are prevented from participating the society in terms of accessing meaningful employment, adequate housing, education, health services and other social determinants of health. The data in the form of quintiles (ordinal scale) is used, where the groups are ranked from least marginalized (level 1) to most marginalized (level 5) [52]. The household dwelling feature measures stability and cohesiveness, considering the types and density of residential population, number of occupants, dwelling ownership, marital status and whether one has moved in the past five years. The material resources take into consideration the accessibility of material needs of individuals and communities in terms of housing, food, clothing, education and employment. The age and labour force feature focus on the impacts of disability and dependence, highlighting individuals without income from employment. Lastly, the racialized and newcomer feature takes into account the proportion of newcomers, or non-white, non-indigenous population that self-identify as a visible minority.

**Table S8:** Continuous features included in the full models.

## 3. Appendix C: Hyperparameters for final LGBM models:

The following are the hyperparameters for the final LGBM model.

|  | Admission Outcome | ED Visit Outcome |
|---|---|---|
| booster | gbdt | gbdt |
| early_stopping_rounds | 17 | 16 |
| learning_rate | -3.827761 | -2.236273 |
| min_data_in_leaf | 32 | 7 |
| max_depth | 7 | 2 |
| num_leaves | 4 | 5 |
| encode | 1 | 1 |
| rebalance | 0 | 0 |

# 4. Appendix D: LLM fine-tuning protocol details:

Our protocol is based on works such as GReaT from Borisov et al. [53] and Pred-LLM from Nguyen et al. [54] The process can be divided into three steps: tabular serialization, fine-tuning process, inference prompting.

## 4.1 Textual encoding or serialization

To leverage from LLM pretraining, we need to transform our tabular data into text. Multiple approaches with various complexities have been explored by Hegselmann et al. [55] and the best performing one was to simply encode each feature value into "*column name* is *feature value,*" and concatenate all features constituting the record, as shown Figure A.1. Missing data is handled by adding "missing" string instead of feature value to keep all column names for each record. A missing feature would be encoded as "*column name* is *missing,*". This method was also implemented in Borisov et al. and Nguyen et al. contributions.

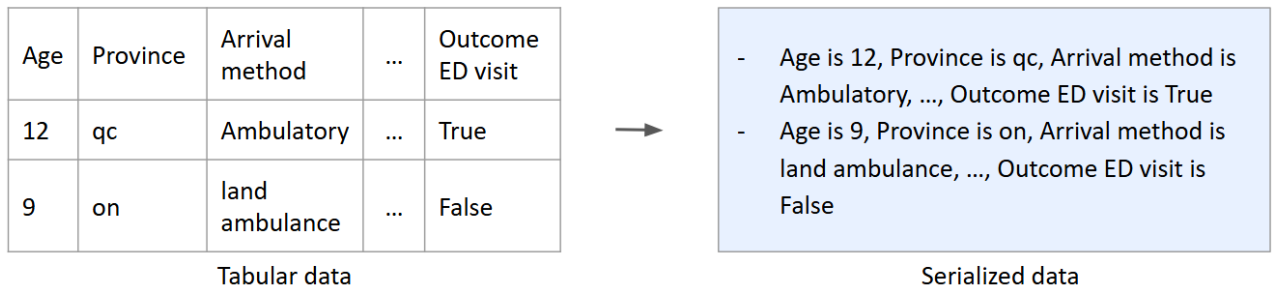

**Figure S1:** Illustration of serialization method.

## 4.2 Fine-tuning process

For the two considered datasets, an outcome variable was always identified: emergency department visit or admission. In such case, we can benefit from the Pred-LLM permutation scheme where the correlation between the outcome variable Y and the other features X is preserved by always placing Y at the end of the sentence and shuffling the rest of the features, as illustrated Figure A.2. Since LLM decoders are trained with masked attention, placing the outcome variables at the end ensures the LLM can attend them through all previously generated features.

The LLMs are then fine-tuned by predicting successive tokens for a certain number of epochs. Following Borisov et al. the number of epochs was set to 200 and we used low-rank adaptation method (LoRA) [56] to efficiently fine-tune the LLM on one NVIDIA A6000 GPU.

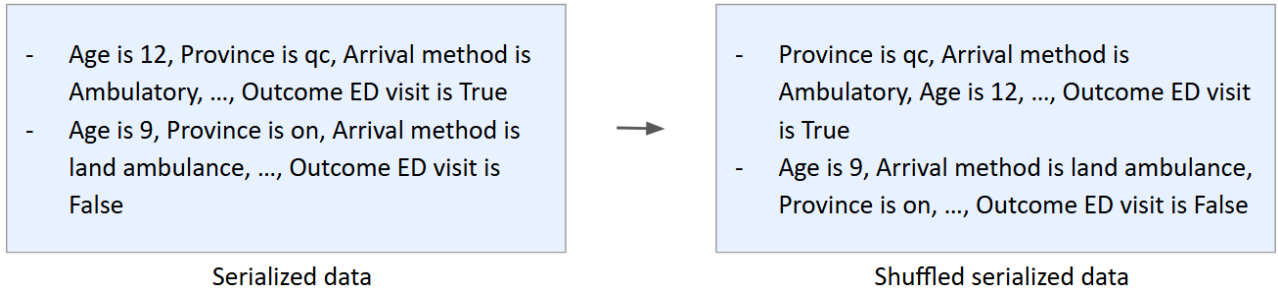

Figure S2: Illustration of features shuffling.

### 4.3 Inference prompting

At inference time, the features get serialized following the textual encoding in Figure A.1 and create the prompt but with the outcome variable missing and the LLM fills in the tokens to produce the answer as shown Figure A.3. The value of the output is then extracted to store the inference outputs as well as prediction scores.

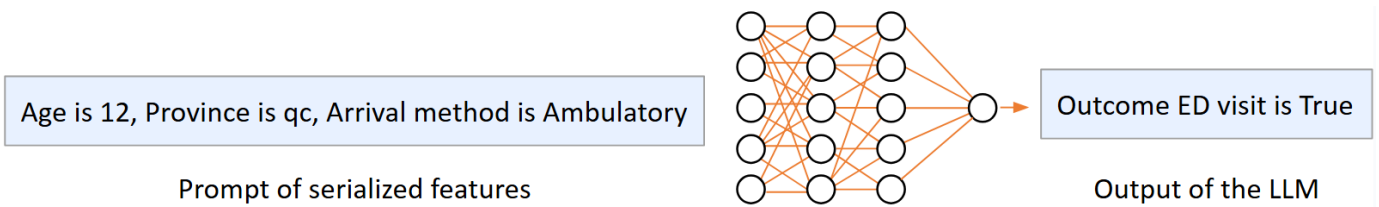

Figure S3: Illustration of inference prompting to perform the classification.

# 5. Appendix E: Naïve Baseline Model Performance

A naïve way to predict high risk patients is to assign all patients as high risk, which maximizes the F1 score. This would give an F1 score of 0.452 and 0.442 for pre- and post-COVID data ED revisits respectively, and 0.212 and 0.302 for hospitalizations respectively. But this is not a meaningful baseline because assigning all patients as high risk would overwhelm the capacity of the institution to provide support.

We therefore considered two random baselines based on a 50% positive prediction (See Table S9) and the prevalence of positive cases (See Table S10). These baselines informed us whether the models are worth deploying.

|  | **Pre-COVID** |  | **Post-COVID** |  |
| --- | --- | --- | --- | --- |
| Existing Baseline Performance | **ED Visit** | **Admission** | **ED Visit** | **Admission** |
| Precision | 0.295 | 0.123 | 0.295 | 0.189 |
| Recall | 0.497 | 0.511 | 0.509 | 0.518 |
| F1 | 0.370 | 0.198 | 0.373 | 0.277 |

**Table S9:** Random assignment of patients based on a 50% positive cases.

|  | **Pre-COVID** |  | **Post-COVID** |  |
| --- | --- | --- | --- | --- |
| Existing Baseline Performance | **ED Visit (30% +ve)** | **Admission (11% +ve)** | **ED Visit (30% +ve)** | **Admission (11% +ve)** |
| Precision | 0.299 | 0.142 | 0.288 | 0.221 |
| Recall | 0.296 | 0.115 | 0.284 | 0.136 |
| F1 | 0.297 | 0.127 | 0.286 | 0.169 |

**Table S10:** Random assignment of patients based on the prevalence of positive cases.

## 6. Appendix F: SHAP Plots for Post-COVID LGBM Models

The following plots were produced for the model trained only on the post-COVID dataset.

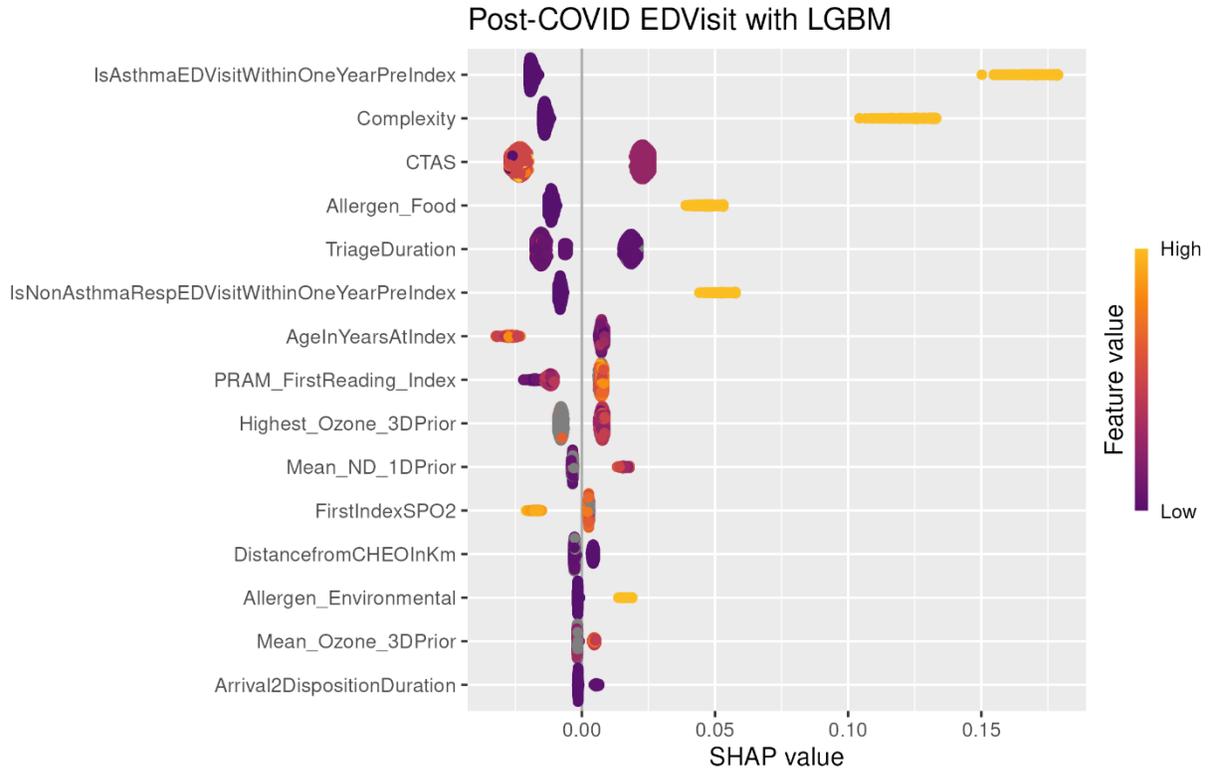

**Figure S4:** SHAP plot for the post-COVID model on the ED Visit outcome.

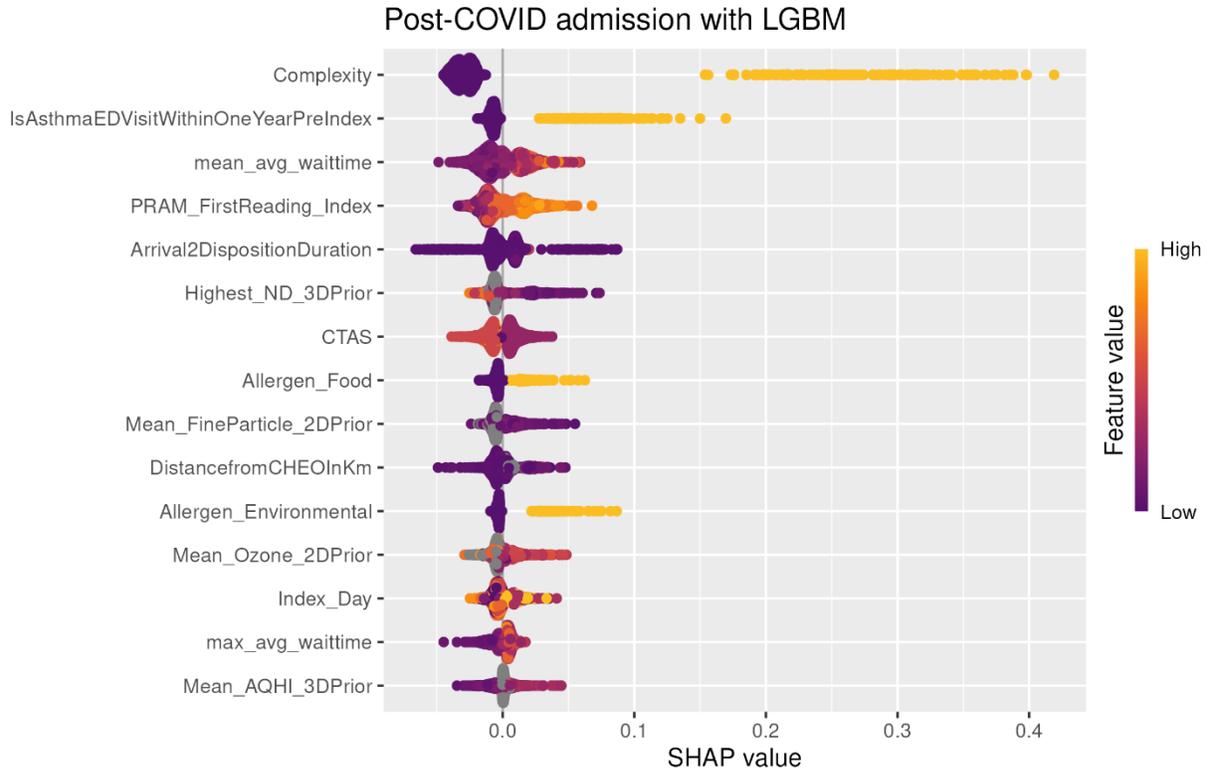

**Figure S5:** SHAP plot for the post-COVID model on the hospital re-admission outcome.

# 7. Appendix G: Consolidated Reporting Guidelines for Prognostic and Diagnostic Machine Learning Models (CREMLS) Checklist

The following is a summary of our reporting according the CREMLS checklist [57,58].

| Study details | |
|---|---|
| 1.1 The medical or clinical task of interest | √ |
| 1.2 The research question | √ |
| 1.3 Current medical or clinical practice | √ |
| 1.4 The known predictors and confounders of what is being predicted or diagnosed | √ |
| 1.5 The overall study design | √ |
| 1.6 The medical institutional settings | √ |
| 1.7 The target patient population | √ |
| 1.8 The intended use of the ML model | √ |
| 1.9 Existing model performance benchmarks for this task | √ |
| 1.10 Ethical and other regulatory approvals obtained | √ |
| The data | |
| 2.1 Inclusion or exclusion criteria for the patient cohort | √ |
| 2.2 Methods of data collection | √ |
| 2.3 Bias introduced due to the method of data collection used | |
| 2.4 Data characteristics | √ |
| 2.5 Methods of data transformation and preprocessing applied | |
| 2.6 Known quality issues with the data | |
| 2.7 Sample size calculation | |
| 2.8 Data availability | √ |
| Methodology | |
| 3.1 Strategies for handling missing data | |
| 3.2 Strategies for addressing class imbalance | |
| 3.3 Strategies for reducing dimensionality of data | |
| 3.4 Strategies for handling outliers | |
| 3.5 Strategies for data augmentation | |
| 3.6 Strategies for model pretraining | √ |
| 3.7 The rationale for selecting the ML algorithm | √ |
| 3.8 The method of evaluating model performance during training | √ |
| 3.9 The method used for hyperparameter tuning | √ |
| 3.10 Model's output adjustments | |
| Evaluation | |
| 4.1 Performance metrics used to evaluate the model | √ |
| 4.2 The cost or consequence of errors | |
| 4.3 The results of internal validation | √ |
| 4.4 The final model hyperparameters | √ |
| 4.5 Model evaluation on an external data set | √ |
| 4.6 Characteristics relevant for detecting data shift and drift | |
| Explainability and transparency | |
| 5.1 The most important features and how they relate to the outcomes | √ |
| 5.2 Plausibility of model outputs | √ |
| 5.3 Interpretation of a model's results by an end user | √ |